\documentclass{article}
\usepackage[utf8]{inputenc}
\usepackage{graphicx,xcolor}
\usepackage{lipsum}
\usepackage{amsmath}
\usepackage{amsfonts}
\usepackage{amssymb}
\usepackage[a4paper, total={6.5in, 10in}]{geometry}
\usepackage{multirow}
\usepackage{float}
\usepackage{comment}
\usepackage{placeins}

\usepackage[backend=biber, style=alphabetic, sorting=ynt]{biblatex}

\author{Mattia Pugliatti\footnote{PhD Student, ESR of Stardust-R, mattia.pugliatti@polimi.it} $ $ and Francesco Topputo\footnote{Full professor}}

\title{{DOORS: Dataset fOr bOuldeRs Segmentation}. Statistical properties and Blender setup}

\date{Politecnico di Milano, Department of Aerospace Science and Technology, Via La Masa 34, 20156, Milan, Italy}

\begin{document}

\maketitle


\begin{abstract}
The capability to detect boulders on the surface of small bodies is beneficial for vision-based applications such as hazard detection during critical operations and navigation. This task is challenging due to the wide assortment of irregular shapes, the characteristics of the boulders population, and the rapid variability in the illumination conditions. Moreover, the lack of publicly available labeled datasets for these applications damps the research about data-driven algorithms. In this work, the authors provide a statistical characterization and setup used for the generation of two datasets about boulders on small bodies that are made publicly available.
\end{abstract}

\textbf{Datasets link:} 10.5281/zenodo.7107409

\section{Introduction}

Data-driven Image Processing (IP) algorithms represent an accurate, robust, generalized alternative to traditional algorithms for vision-based applications about small bodies. They can either be used for navigation applications as well as or for IP tasks that enable on-board autonomous capabilities.

The lack of publicly available labeled datasets (both synthetic and from real missions) is a critical showstopper for the development of these types of algorithms \cite{song2022deep}. In their absence, the algorithm designer has two choices: either to generate one on its own or to use unsupervised learning for the task. In the former case, it requires interdisciplinary skills that also cover aspects related to the capability to perform realistic renderings in an artificial environment. In the latter, it largely reduces the design space of the algorithms. 

However, the design of a dataset generator is not a simple task and requires a non-negligible effort that may shift the focus from the algorithm design. Moreover, as the sole possession of artificial dataset generators poses a strategic advantage both in industrial and research applications, they are often not publicly available. The same reasoning applies also to the datasets. 

\begin{figure}[ht]
	\centering
	\includegraphics[width=0.7\textwidth]{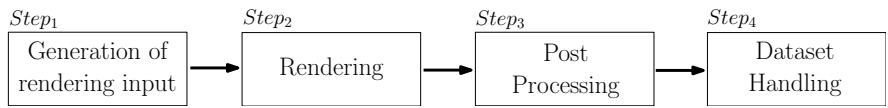}
	\caption{Pipeline used for the datasets generation.}
	\label{fig:Dataset_generation_strategy}
\end{figure}

In this work, artificial environments are specifically designed to generate large amounts of synthetic labeled images with boulders on small bodies. To do so, Blender\footnote{https://www.blender.org/, retrieved 13th of September, 2022.} is used due to its simplicity, extensive prior usage, large support community, and open-source licensing. Using these artificial environments, two datasets are generated, for simplicity referred to as $DS_1$, and $DS_2$. These datasets have been originally designed for the work presented in \cite{pugliatti2022_siw}, but their application can be extended to additional cases. This paper presents the Blender setup used to generate them as well as their statistical properties. The datasets are available for download at \cite{DatasetZenodoBoulders}. The 4-steps pipeline that has been used for their generations is illustrated in Figure \ref{fig:Dataset_generation_strategy}. 

First, the rendering input needed by the artificial environment is generated as a \textit{txt} file, which is read by a python script in Blender that loops through the input labels that controls all the objects in the artificial environment. After all image-mask pairs are generated, they are further processed in \textit{Step}$_3$ and then prepared in a dataset-ready format to be used for training data-driven algorithms. 
\FloatBarrier

\section{Blender setup of $\mathbf{DS_1}$}

The artificial environment in Blender that is used to generate $DS_1$ is made by 4 elements: 

\begin{enumerate}
    \item A single randomly-generated boulder whose Center of Mass (CoM) is positioned in the center of the Blender reference frame
    \item A unitary spherical mesh made of 16'258 vertexes and 32'512 faces
    \item A camera, modeled with a $256\times256$ px size sensor and a FOV of $10\times10$ deg
    \item A Sun lamp illuminating the scene
\end{enumerate}

These elements are then positioned within the environment accordingly to a \textit{txt} file that is used as an input file. The input files are composed of $N$ rows and $16$ columns, where $N$ is the number of image-mask pairs as samples that will be generated. The meaning of each column is represented in Table \ref{tab:labels_DS1}, while the names of the splits used in this work for the training, validation, and test sets are illustrated in Table \ref{tab:DS1_splits_names}

\begin{table}[ht]
    \centering
    \begin{tabular}{rccl}
        \hline
        \hline
        Col \# & Units & Symbol & Name \\
        \hline
        1,2,3 & BU & $X,Y,Z$ & Camera position \\
        4,5,6 & BU & $X_S,Y_S,Z_S$ & Sun direction vector \\
        7 & deg & $\theta_b$ & Camera boresight rotation \\
        \hline
        8 & - & $ID$ & ID of the boulder \\
        9 & BU & $s$ & Scale of the boulder \\
        10 & - & $a_s$ & Albedo of the surface \\
        11 & - & $a_b$ & Albedo of the boulder \\
        \hline
        12 & - & $I$ & Sun's intensity \\
        \hline
        13 & - & $\nu_{s1}$ & Scale of the noise pattern \\
        14 & - & $\nu_r$ & Roughness of the surface \\
        15 & - & $\nu_d$ & Distortion of the surface \\
        16 & - & $\nu_{s2}$ & Scale of the surface displacement \\
        \hline
    \end{tabular}
    \caption{Labels used in $DS_1$ to generate the renderings.}
    \label{tab:labels_DS1}
\end{table}

\begin{table}[ht]
    \centering
    \begin{tabular}{c l c}
        \hline
        \hline
        Split & Name &  $N$ \\
        \hline
         $Tr$ & \textit{T\_30000\_b\_2022-08-02\;11.14.22.txt} & 30181 \\
         $V$ & \textit{V\_5000\_b\_2022-08-02\;11.15.52.txt} & 5044 \\
         $Te_1$ & \textit{Te1\_5000\_b\_2022-08-02\;11.16.00.txt} & 5044 \\
         $Te_2$ & \textit{Te2\_5000\_ub\_2022-08-02\;11.16.11.txt} & 5000 \\
        \hline
    \end{tabular}
    \caption{Input files for the renderings of $DS_1$.}
    \label{tab:DS1_splits_names}
\end{table}

As it is possible to see in Table \ref{tab:DS1_splits_names}, the number of samples $N$ is not reflected by the number used in the \textit{txt} name. This is because, at the beginning of the input file generation, a rough number is expressed (e.g. 30'000), which may not be the final number of samples contained in the \textit{txt}. This discrepancy is due to the balancing of the dataset. For example, the $Tr$, $V$, and $Te_1$ splits are balanced ones and as a result thus they have slightly more samples than the ones inputted at the beginning of the procedure. On the other hand, $Te_2$ has the same number, due to its unbalanced nature. The balancing regards the distribution of the phase angle $\Psi$ among the samples, as will be illustrated later on. The balancing is handled by a flag $\nu_{bal}$. Depending on this flag the number of points to be generated will respectively be $N'= 5M$ for the balanced case, and $N' = M$ for the unbalanced one, where $M$ represents the rough number of desired samples expressed at the beginning of the procedure (e.g. 30'000 or 5000). 

A cloud of $N'$ points is therefore generated in spherical coordinates such that the range, equatorial and elevation angles are distributed uniformly as $\rho \in [2.0860,10.4301]$ BU\footnote{Blender Units (BU)}, $\theta \in [10,170]$ deg, $\phi \in [-30,30]$ deg. These coordinates are then transformed into cartesian ones and represent camera positions in the Blender reference frame. The attitude of the camera at each position is set with the primary pointing towards the center of the Blender reference frame and with the secondary pointing commanded by a uniform random rotation about the boresight $\theta_b \in [0,360]$ deg. Each pose is associated with a unique image-masks set, that represents a dataset sample.

\begin{figure}[ht]
	\centering
	\includegraphics[width=0.6\textwidth]{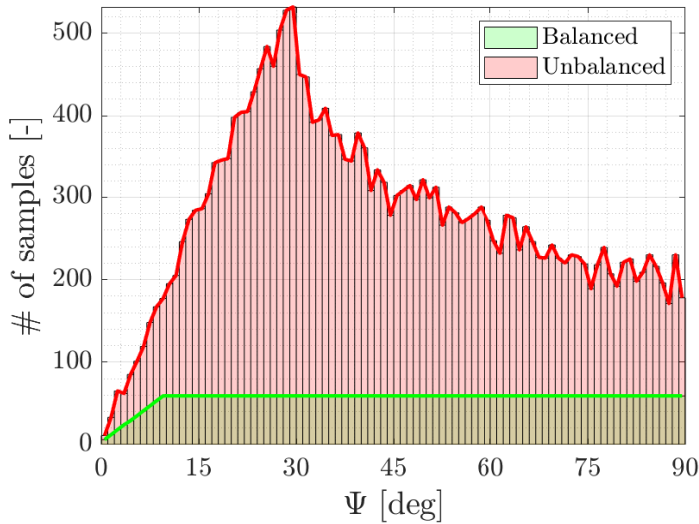}
	\caption{Example between a balanced and unbalanced distributions respectively of 5044 and 25000 samples.}
	\label{fig:balanced_example}
\end{figure}

The illumination conditions are defined by the orientation and intensity of the Sun's lamp in Blender. The intensity is handled by a random uniform value $I \in [35,65]$. For what concern the orientation, the light is constrained to be directed from the equatorial plane of the Blender reference frame. The angle between the camera position's vector projection on the equatorial plane and a line-of-sight vector on the equatorial plane generated by a random uniformly generated angle $\theta_S \in[-90,90]$ is computed. Whenever this angle is between $0$ deg and $180$ deg it is considered viable from a visibility and illumination point of view. The phase angle $\Psi$ between the camera position vector and the illumination vector is computed for each sample and pruning is performed based on the balance flag $\nu_{bal}$. In the case in which a balanced dataset is desired, the distribution of the phase angle is forced to assume a linear increase up to $10$ deg, after which a constant value is kept from $10$ deg to $90$ deg, as illustrated in the example in Figure \ref{fig:balanced_example}. The balanced distribution is set to have a number of samples as close as possible to the target value, by selecting subsets from the larger unbalanced distribution. 

\begin{figure}[ht]
	\centering
	\includegraphics[width=1\textwidth]{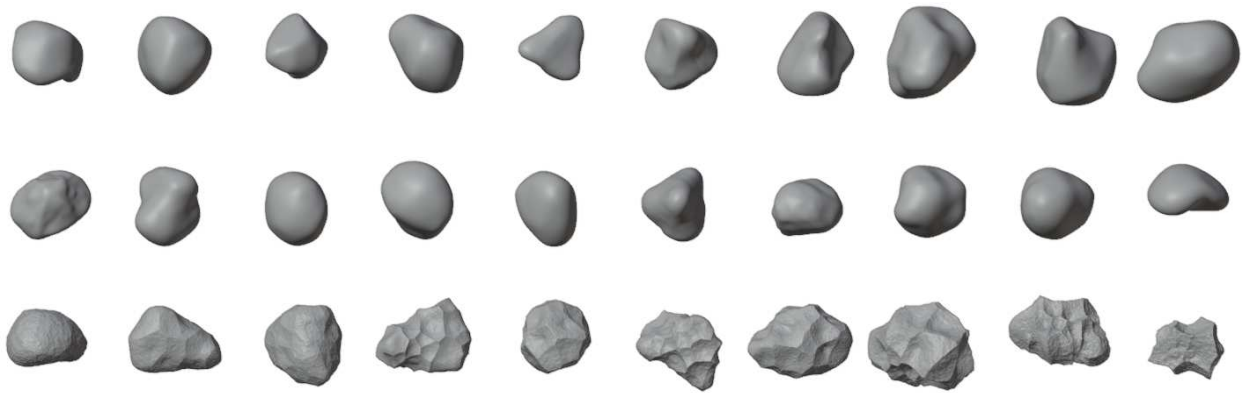}
	\caption{The 30 archetype shapes that represent single instances of boulders in $DS_1$. From top to bottom the \textit{ice}, \textit{river}, and \textit{asteroid} classes are shown.}
	\label{fig:BouldersSample}
\end{figure}

Using the \textit{Rock Generator} add-on in Blender, a set of 30 boulder archetype shapes is generated, as illustrated in Figure \ref{fig:BouldersSample}. These are divided into three classes defined within the \textit{Rock Generator} add-on, characterized by different default settings. These are the \textit{ice}, \textit{river}, \textit{asteroid} classes. 

For each sample, one of the 30 boulders is randomly selected with a uniform random sample $ID \in[0,29]$ and positioned in the origin of the reference frame. To further increase variability, each boulder is also randomly scaled with a uniform random value $s$, which is selected according to the specific class. 

\begin{figure}[ht]
	\centering
	\includegraphics[width=0.55\textwidth]{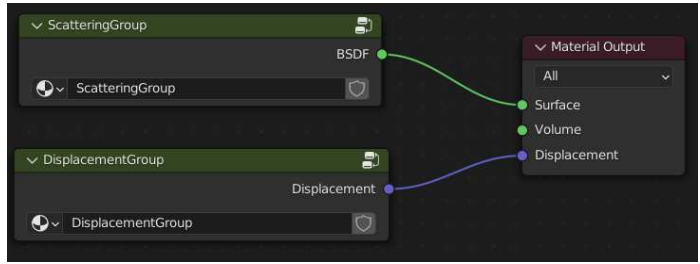}
	\caption{View of the shading tab in Blender regarding the surface mesh.}
	\label{fig:surface_variation}
\end{figure}

In the shading tab, exploiting the Open Shading Language (OSL) package in Blender, the Akimov law is used to simulate ligth scattering. The modifications in the shading tab to perform scattering law corrections have been performed following the procedure illustrated in \cite{pentila2021scattering}. The albedo of the surface is selected from a random uniform distribution as $a_s \in [0.05, 0.15]$. The albedo of each boulder $a_b$ is then generated by multiplying a random uniform adimensional coefficient $a_b' \in [0.7,3]$ with the corresponding $a_s$ for each sample. This is done to guarantee variability between surface and boulder's albedo through the datasets. While both the boulders and surface are simulated through scattering laws in the shading tab, the spherical mesh representing the surface is also modified to generate a variety of different roughness surroundings, as illustrated in Figure \ref{fig:surface_variation}.

\begin{figure}[ht]
    \centering
    \includegraphics[width=0.55\textwidth]{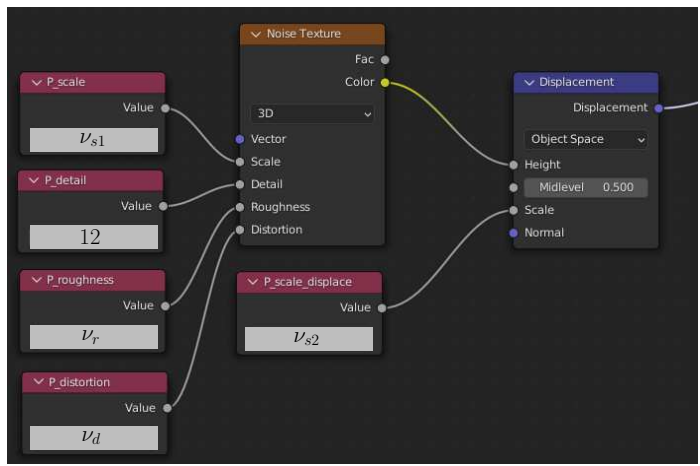}
    \caption{View within the \textit{DisplacementGroup} of the shading tab in Blender.}
    \label{fig:surface_displacement}
\end{figure}

The settings governing the surface displacement are the \textit{Scale}, \textit{Detail}, \textit{Roughness}, \textit{Distortion} settings within the \textit{Noise Texture} block and the \textit{Scale} setting within the \textit{Displacement} block, as illustrated in Figure \ref{fig:surface_displacement}. These are varied as $\nu_{s1} \in [4,6]$, $\nu_r \in [0.4,0.6]$, $\nu_d \in [0.05,0.2]$, $\nu_{s2} \in [-0.25,0.25]$. 

\begin{figure}[ht]
	\centering
	\includegraphics[width=1\textwidth]{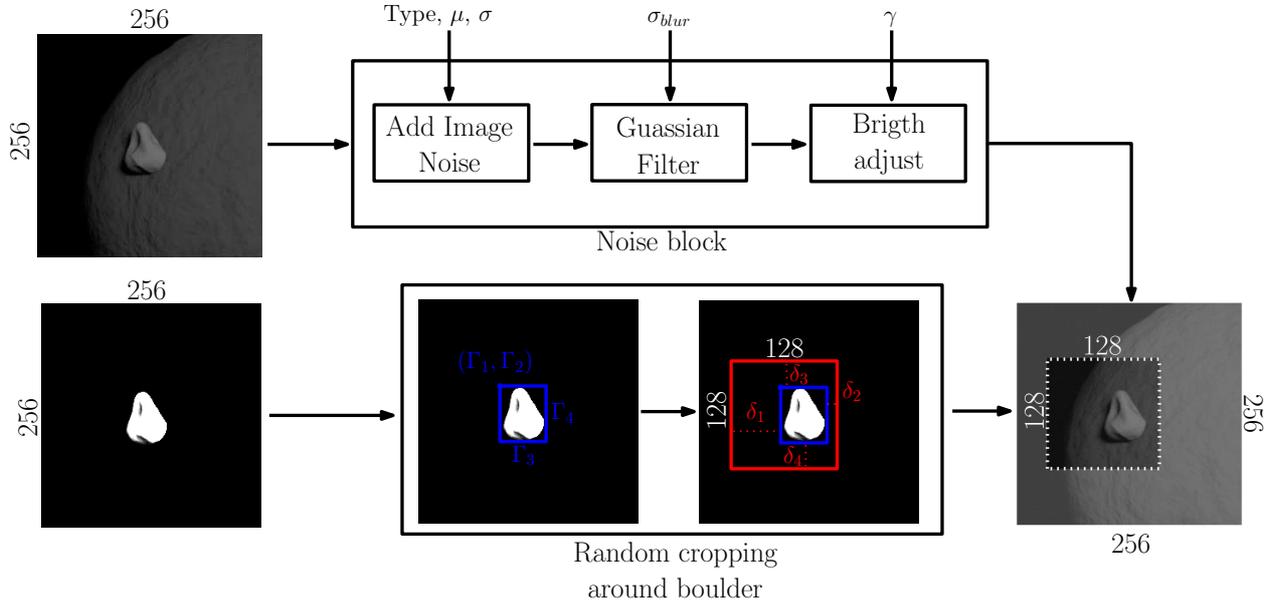}
	\caption{Post-processing pipeline used in $DS_1$. For this dataset $\text{Type}= Gaussian$, $\mu=0.1$, $\sigma = 0.0001$, $\sigma_{blur}=0.5$, and $\gamma=1.2$.}
	\label{fig:PPP_DS1}
\end{figure}

Following this procedure, renderings are executed looping through the different rows of each input file. By setting different pass indices to the surface and boulders and using \textit{Cycles} as a rendering engine, its ray-tracing capabilities are used to obtain not only the grayscale images but their corresponding ground truth masks about surface and boulders. By combining this mask with the illumination conditions it is also possible to obtain the masks with shadows. Both sets of masks are generated for boulder and surface layers. During rendering, image-masks sets are rendered at $256\times256$ resolutions. However, a post-processing pipeline is put in place to perform random cropping and artificial noise addition. This pipeline is illustrated in Figure \ref{fig:PPP_DS1}.

\begin{figure}[ht]
	\centering
	\includegraphics[width=0.8\textwidth]{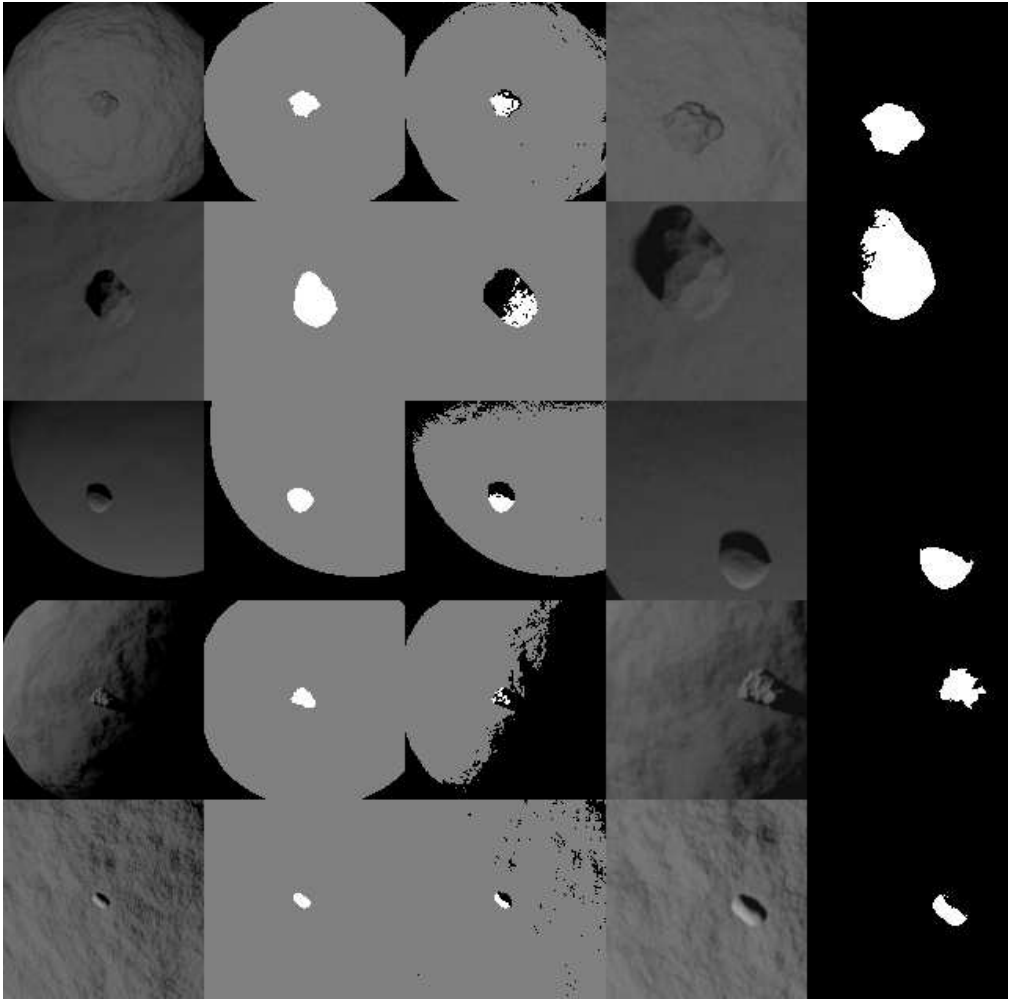}
	\caption{Sample of image-mask sets of $DS_1$. From left to right: $256\times256$ grayscale renderings in Blender, masks without shadows, masks with shadows, followed by $128\times128$ noisy and randomly cropped grayscale images, and relative masks with shadows.}
	\label{fig:DS1_sample}
\end{figure}

Artificial noise is added to images, while blobs analysis is performed on the boulder's mask to perform randomized cropping which achieves a twofold objective: to reduce the image resolution to $128\times128$ px, and to scatter the boulder's across the entire image plane, the latter being fundamental for generalization. It is noted that noise is added merely to reduce the possible domain gap between synthetic and real imagery and not to represent a specific camera model. The list of variables saved after post-processing is illustrated in Table \ref{tab:labels_DS1_1}. Note that these variables are saved in \textit{mat} files with the corresponding names of the splits. Finally a sample of image-mask sets of $DS_1$ is illustrated in Figure \ref{fig:DS1_sample}.

\begin{table}[ht]
    \centering
    \begin{tabular}{rccl}
        \hline
        \hline
        Col \# & Units & Symbol & Name \\
        \hline
        1,2 & px & $CoB_{u,v}$ & CoB coordinates in the rendered images \\
        3,4 & px & $CoF_{u,v}$ & CoF coordinates in the rendered images \\
        5,6 & px & $CoB_{u,v}$ & CoB coordinates in the post-processed images \\
        7,8 & px & $CoF_{u,v}$ & CoF coordinates in the post-processed images \\        
        9,10,11,12 & px & $\Gamma$ & Components of the bounding box around the boulder mask with shadow \\
        13,14,15,16 & px & $\delta$ & Padding values to reach target size image \\
        17 & px & $\delta^{T}$ & Target size of the image \\
        18 & - & $N_{blobs}$ & Number of blobs detected in the image \\
        \hline
    \end{tabular}
    \caption{Labels generated after post-processing in $DS_1$ images.}
    \label{tab:labels_DS1_1}
\end{table}

\FloatBarrier
\section{Blender setup of $DS_2$}

The procedure adopted to generate the image-label pairs of $DS_2$ shares the same steps as the one used for $DS_1$ with few important differences. First of all, the Blender environment is represented by 4 elements: 

\begin{enumerate}
    \item A medium resolution mesh of the (65803) Didymos asteroid made of 652'032 faces that represents the surface
    \item A particle system which scatters randomized populations of boulders from the collection of 30 samples in Figure \ref{fig:BouldersSample} across the surface
    \item A camera, modeled with a $128\times128$ px size sensor and a FOV of $10\times10$ deg
    \item A Sun lamp illuminating the scene
\end{enumerate}

The procedure to generate the input \textit{txt} to use in Blender is the same as the one illustrated for $DS_1$. This time, however, only $10$ of the original $16$ labels are used and an additional one representing the random rotation of the asteroid around its axis is introduced as $\theta_{as}$. Also, the $\rho$ has been generated in a different interval, $\rho \in [3.9,13]$ BU. These labels are illustrated in Table \ref{tab:labels_DS2} while the name of the input file is summarized in Table \ref{tab:DS2_splits_names}. Note that for simplicity the columns of the labels are kept consistent with the ones in $DS_1$ and that the unnecessary columns are simply skipped during the rendering phase. 

\begin{table}[ht]
    \centering
    \begin{tabular}{rrrl}
        \hline
        \hline
        Col \# & Units & Symbol & Name \\
        \hline
        1,2,3 & BU & $X,Y,Z$ & Camera position \\
        4,5,6 & BU & $X_S,Y_S,Z_S$ & Sun direction vector \\
        7 & deg & $\theta_{b}$ &Camera boresight rotation \\
        \hline
        8 & deg & $\theta_{as}$ & Angular rotation of the small body\\
        10 & - & $a_s$ & Albedo of the surface \\
        11 & - & $a_b$ & Albedo of the boulder \\
        \hline
        12 & - & $I$ & Sun's intensity \\
        \hline
    \end{tabular}
    \caption{Labels used in $DS_2$ to generate the renderings.}
    \label{tab:labels_DS2}
\end{table}

\begin{table}[ht]
    \centering
    \begin{tabular}{c l c}
        \hline
        \hline
        Split & Name & $N$ \\
        \hline
         $Tr$ & \textit{20000\_b\_2022-09-13\;22.39.08.txt} & 20095 \\
         $V$ & \textit{5000\_b\_2022-09-13\;22.40.10.txt} & 5044 \\
         $Te_1$ & \textit{5000\_b\_2022-09-13\;22.40.14.txt} & 5044 \\
         $Te_2$ & \textit{5000\_ub\_2022-09-13\;22.40.20.txt} & 5000 \\
        \hline
    \end{tabular}
    \caption{Input files for the renderings of $DS_2$.}
    \label{tab:DS2_splits_names}
\end{table}

The boulder population in $DS_2$ is handled by the particle system with the settings illustrated in Table \ref{tab:BouldersPopulation}, divided by the classes of the \textit{rock generator} add-on. 

\begin{table}[ht]
    \centering
    \begin{tabular}{r c c c}
        \hline
        \hline
        Size & Ice & River & Asteroid \\
        \hline
        Small & 2500 & 2500 & 2500 \\
        Medium & 200 & 200 & 200 \\
        Large & 2 & 2 & 2 \\
        \hline
    \end{tabular}
    \caption{Settings of the particle system in Blender for $DS_2$.}
    \label{tab:BouldersPopulation}
\end{table}

\begin{figure}[ht]
	\centering
	\includegraphics[width=1\textwidth]{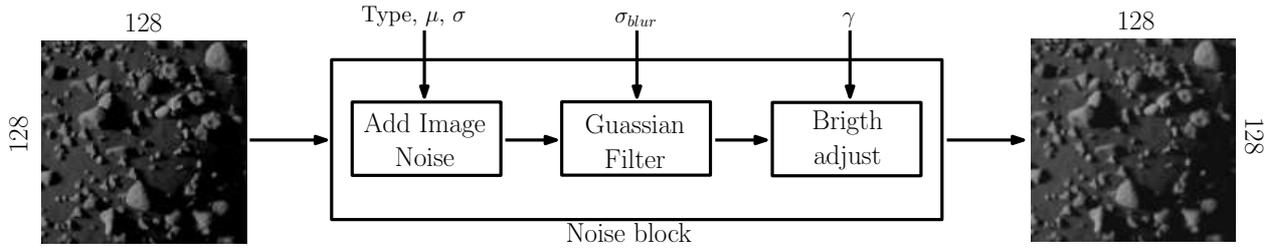} 
	\caption{Post-processing pipeline used in $DS_2$. For this dataset $\text{Type}= Gaussian$, $\mu=0.1$, $\sigma = 0.0001$, $\sigma_{blur}=0.5$, and $\gamma=1.2$.}
	\label{fig:PostProcessDS2}
\end{figure}

\begin{figure}[ht]
	\centering
	\includegraphics[width=0.4\textwidth]{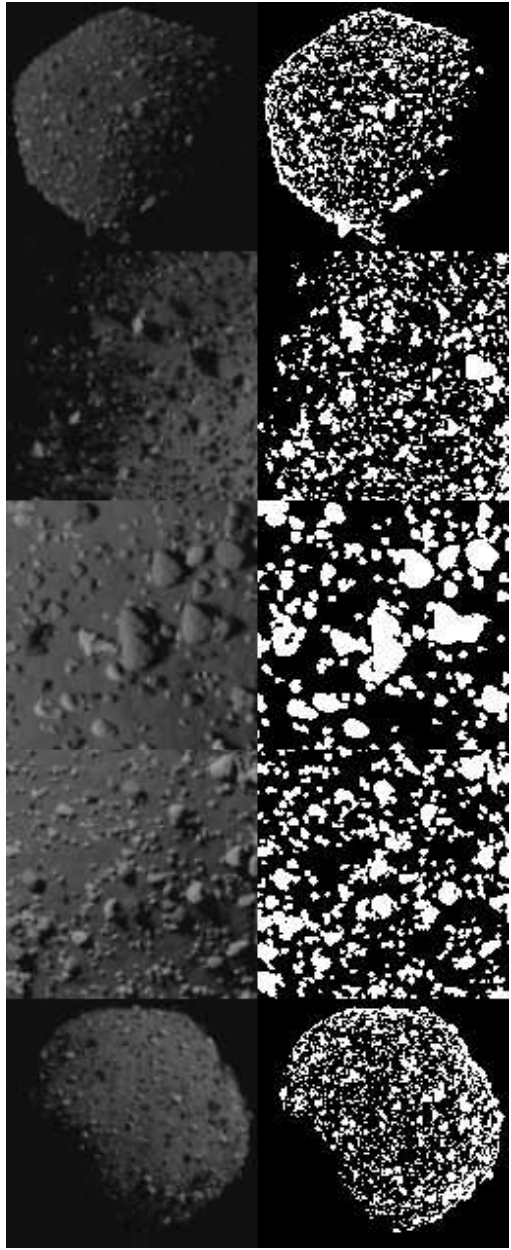}
	\caption{Sample of image-label pairs of $DS_2$. $128\times128$ noisy grayscale images (left) and relative boulder masks (right).}
	\label{fig:DS2_sample}
\end{figure}

Scattering laws are also applied in the shading tab as in the case of $DS_1$, note that this time, however, further randomization is added to the albedo of every single boulder by modifying the input value from the \textit{txt} as follows:  

\begin{equation}
    a_b'=max
    \left\{\begin{matrix}
    a_{b} + rand(-0.5,0.5)\sigma_b
    \\
    0.03
    \end{matrix}\right.
\end{equation}
where $\sigma_b$ is an arbitrary parameter set to $0.35$. After rendering, only artificial noise is added to images, as illustrated by the post-processing pipeline of $DS_2$ in Figure \ref{fig:PostProcessDS2}. A sample of image-mask sets from $DS_2$ is illustrated in Figure \ref{fig:DS2_sample}.

\FloatBarrier
\section{Statistical properties of $DS_1$}

The $DS_1$ dataset has been designed for different IP applications. Among these, the authors recognize boulders identification, segmentation, centroid regression, and navigation. The main statistical properties of $DS_1$ are illustrated in detail. Note that all histograms have been plotted with the value of the relative probability on the y-axis. 

\begin{figure}[ht]
	\centering
	\includegraphics[width=1.0\textwidth]{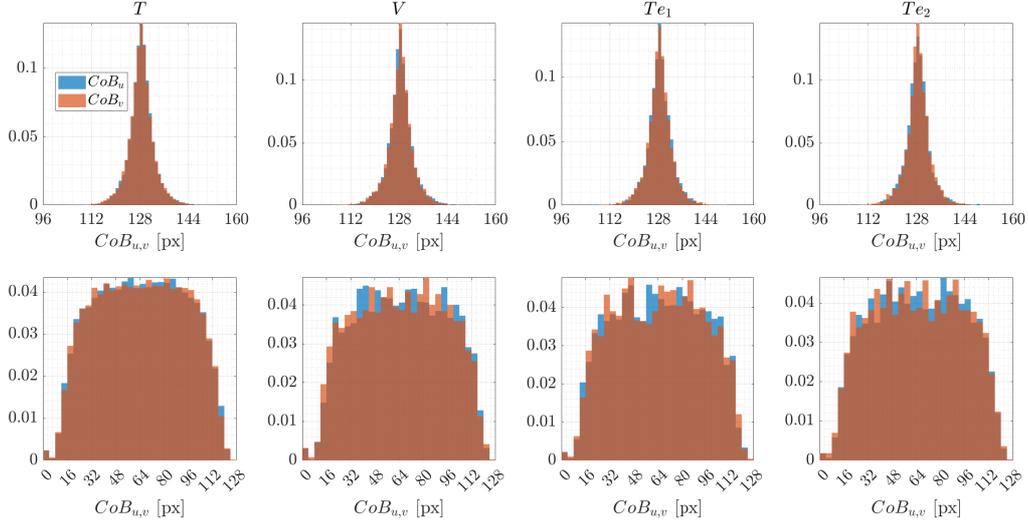}
	\caption{Histograms of the CoB coordinate of the boulders in the $256\times256$ (top) and $128\times128$ px (bottom) images of $DS_1$. Bin-width used is 1 (top) and 4 (bottom) px.}
	\label{fig:reg_histograms_1}
\end{figure}

\begin{figure}[ht]
    \begin{minipage}{.5\linewidth}
        \centering
        \includegraphics[scale=.5]{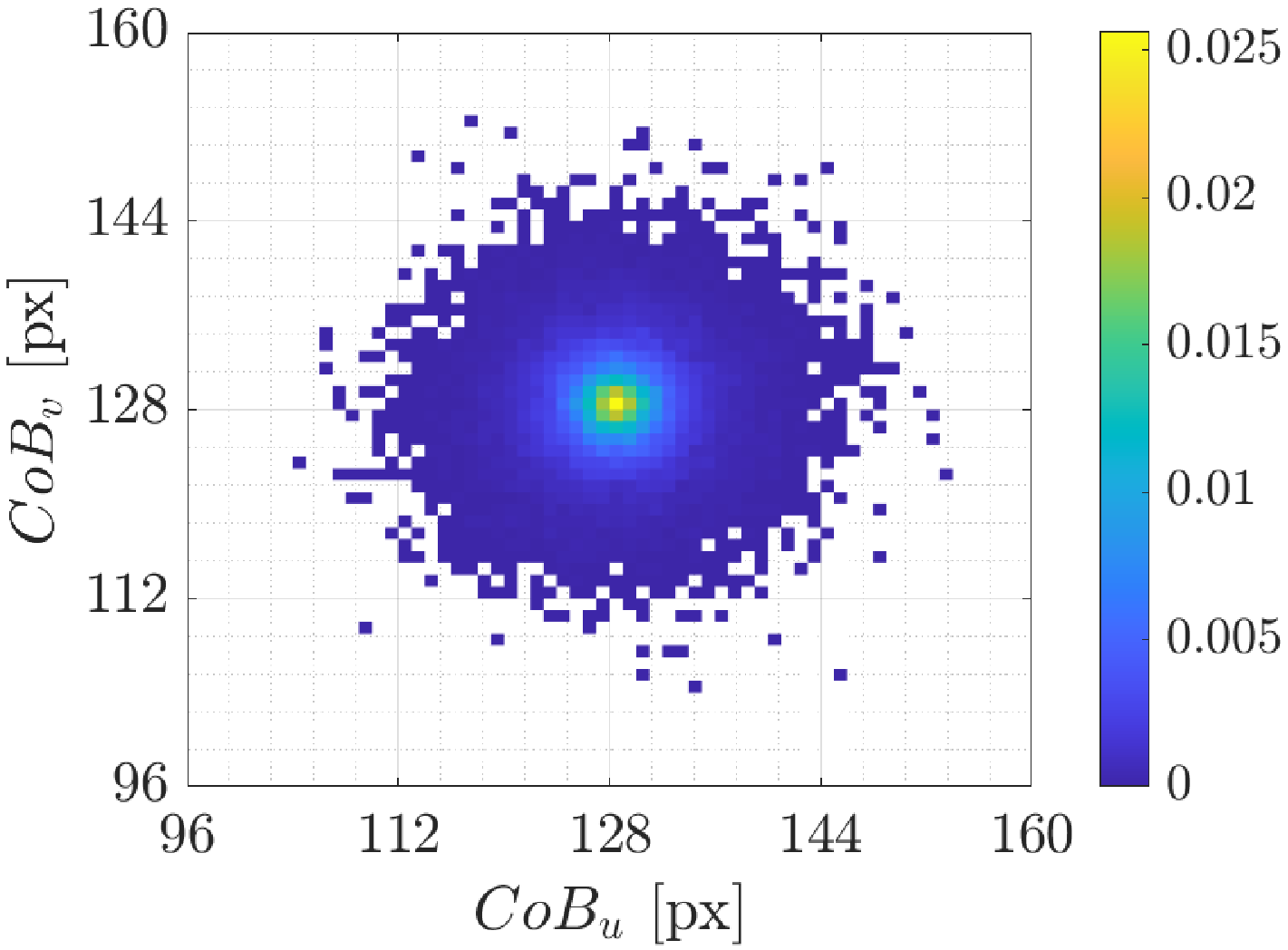}
    \end{minipage}%
    \begin{minipage}{.5\linewidth}
        \centering
        \includegraphics[scale=.5]{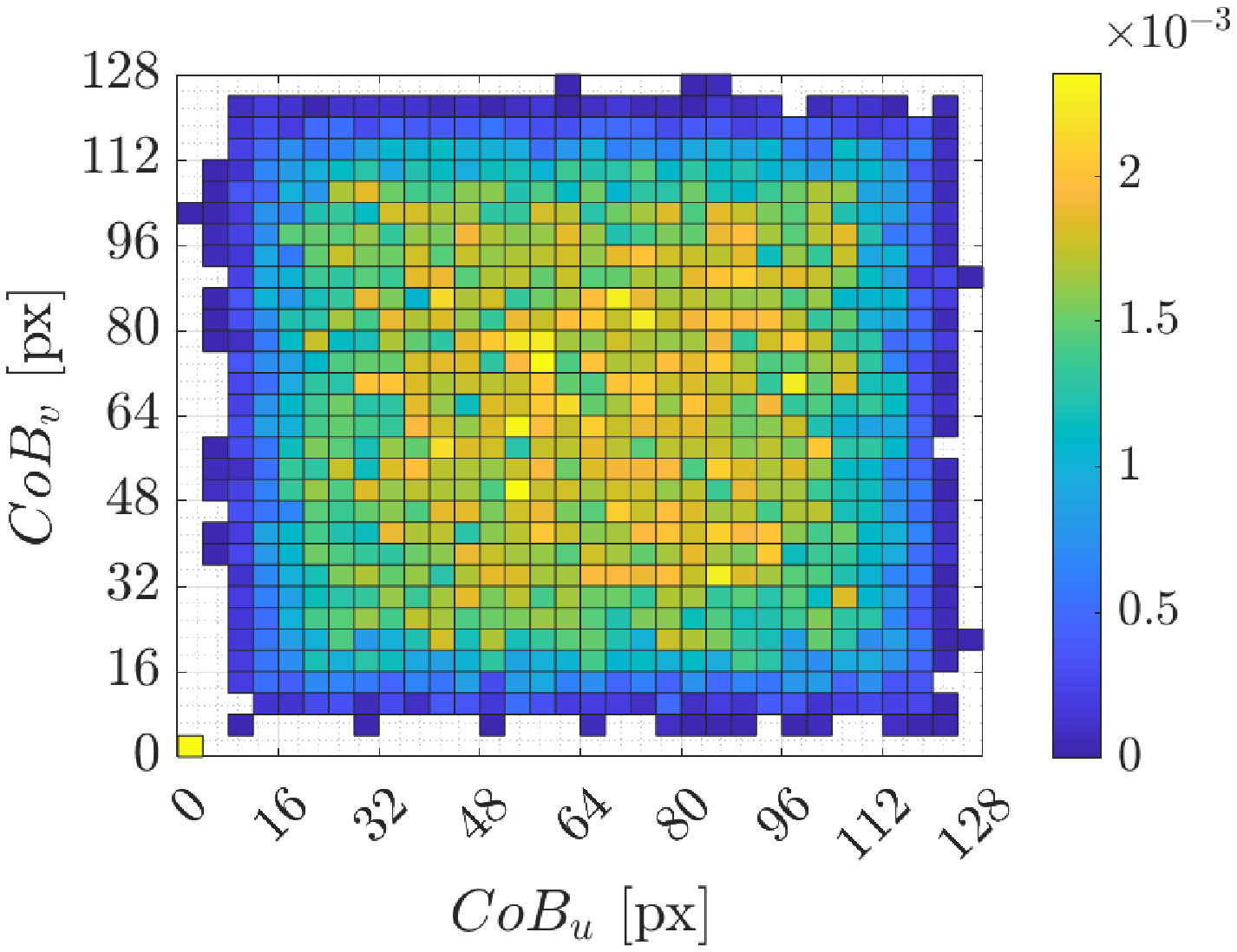}
    \end{minipage}\par\medskip
\caption{2D histograms of the CoB coordinates of the boulders in the $256\times256$ (left) and $128\times128$ px (right) images of $DS_1$. Bin-width used is 1 (left) and 4 (right) px.}
\label{fig:reg_histograms_23}
\end{figure}

Figure \ref{fig:reg_histograms_1} represents the distributions of the CoB coordinates of each boulder across the different splits of $DS_1$, while Figure \ref{fig:reg_histograms_23} represents the same distributions in 2D plots. From these figures is possible to appreciate the beneficial effects of the post-processing in spreading the CoB coordinates uniformly across the image plane. 

\begin{figure}[ht]
    \begin{minipage}{.4\linewidth}
        \centering
        \includegraphics[scale=.4]{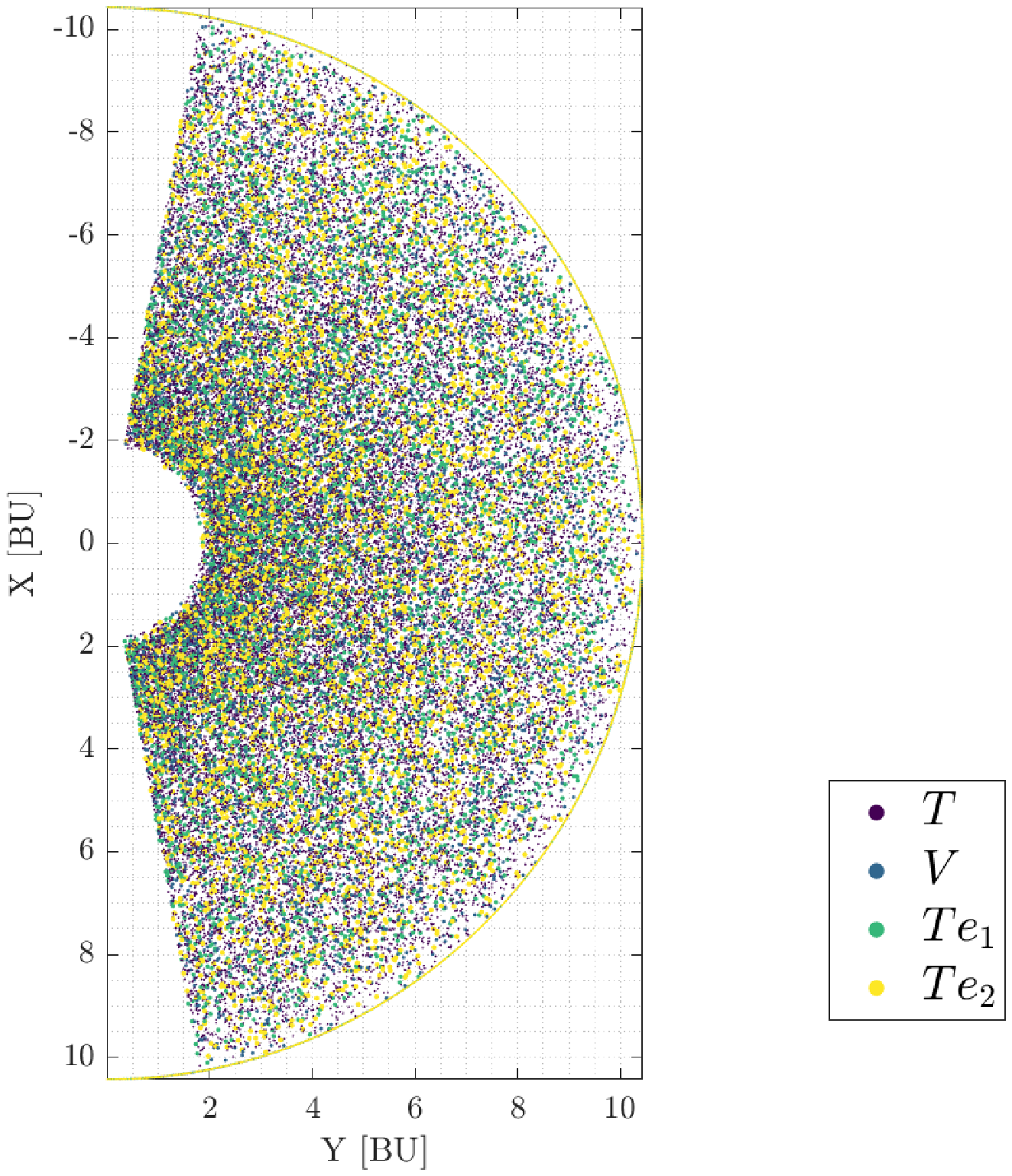}
    \end{minipage}%
    \begin{minipage}{.4\linewidth}
        \centering
        \includegraphics[scale=.4]{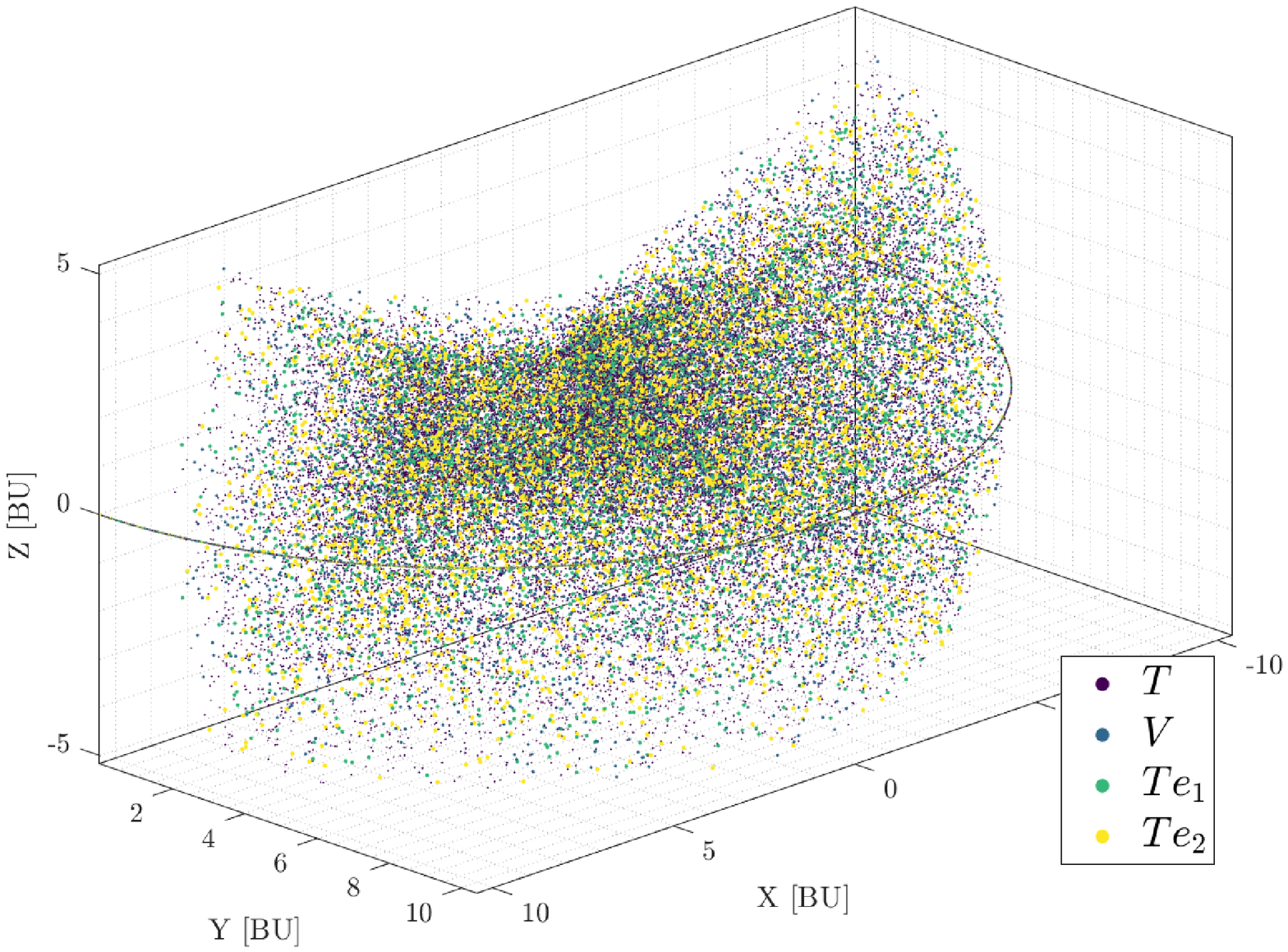}
    \end{minipage}\par\medskip
\caption{Cloud of points of positions and Sun's direction used in the various splits of $DS_1$ in the Blender reference frame. Sun direction varies only in the equatorial plane.}
\label{fig:DS1_XYZ}
\end{figure}

\begin{figure}[ht]
	\centering
	\includegraphics[width=1.0\textwidth]{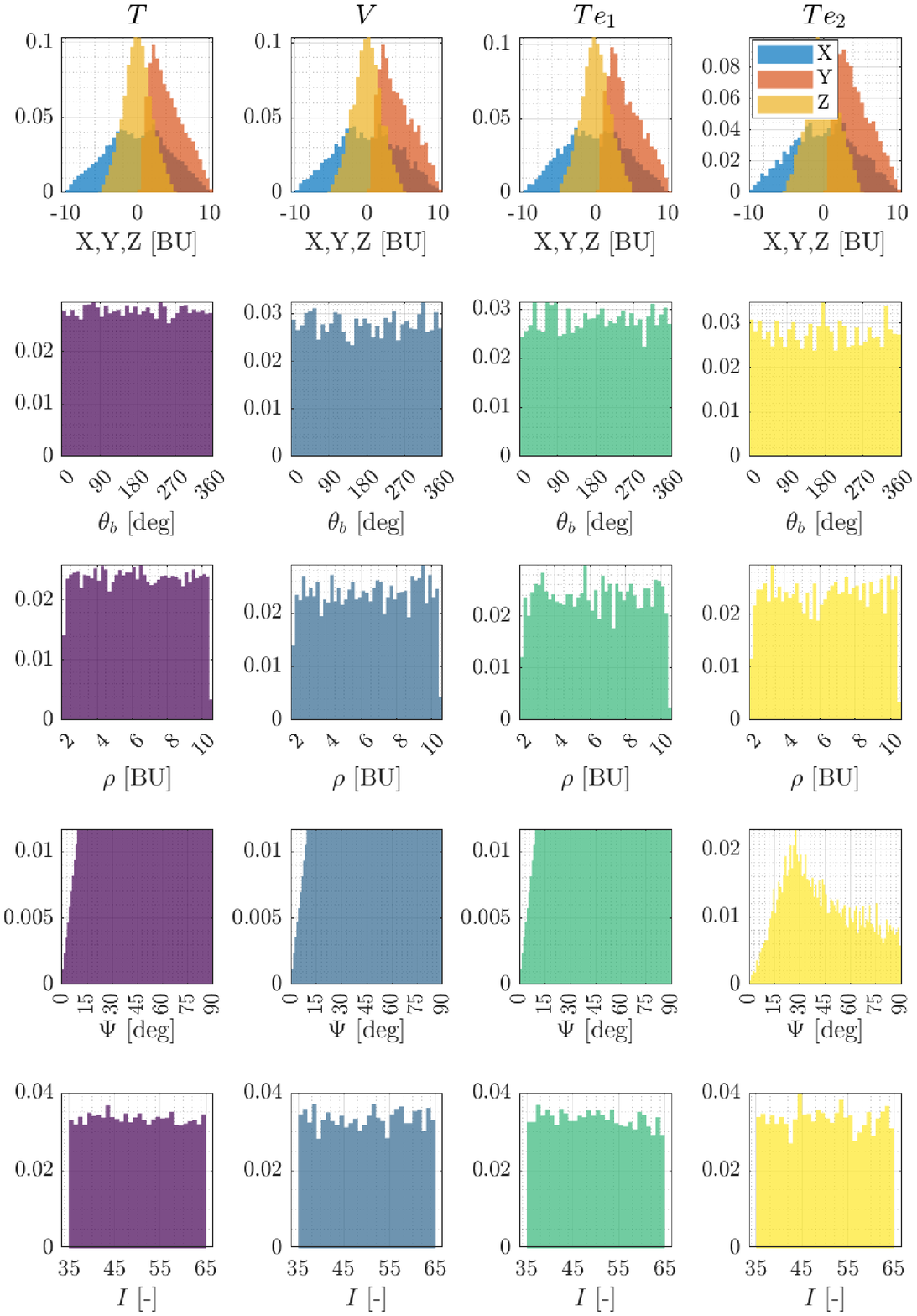}
	\caption{Histograms of properties of the various splits of $DS_1$. From top to bottom: Cartesian coordinates, boresight rotation, range, phase angle, and Sun's intensity. Bin-width used from top to bottom: 0.5, 10 deg, 0.2, 1 deg, 1.}
	\label{fig:DS1_histograms}
\end{figure}

\begin{figure}[ht]
	\centering
	\includegraphics[width=0.7\textwidth]{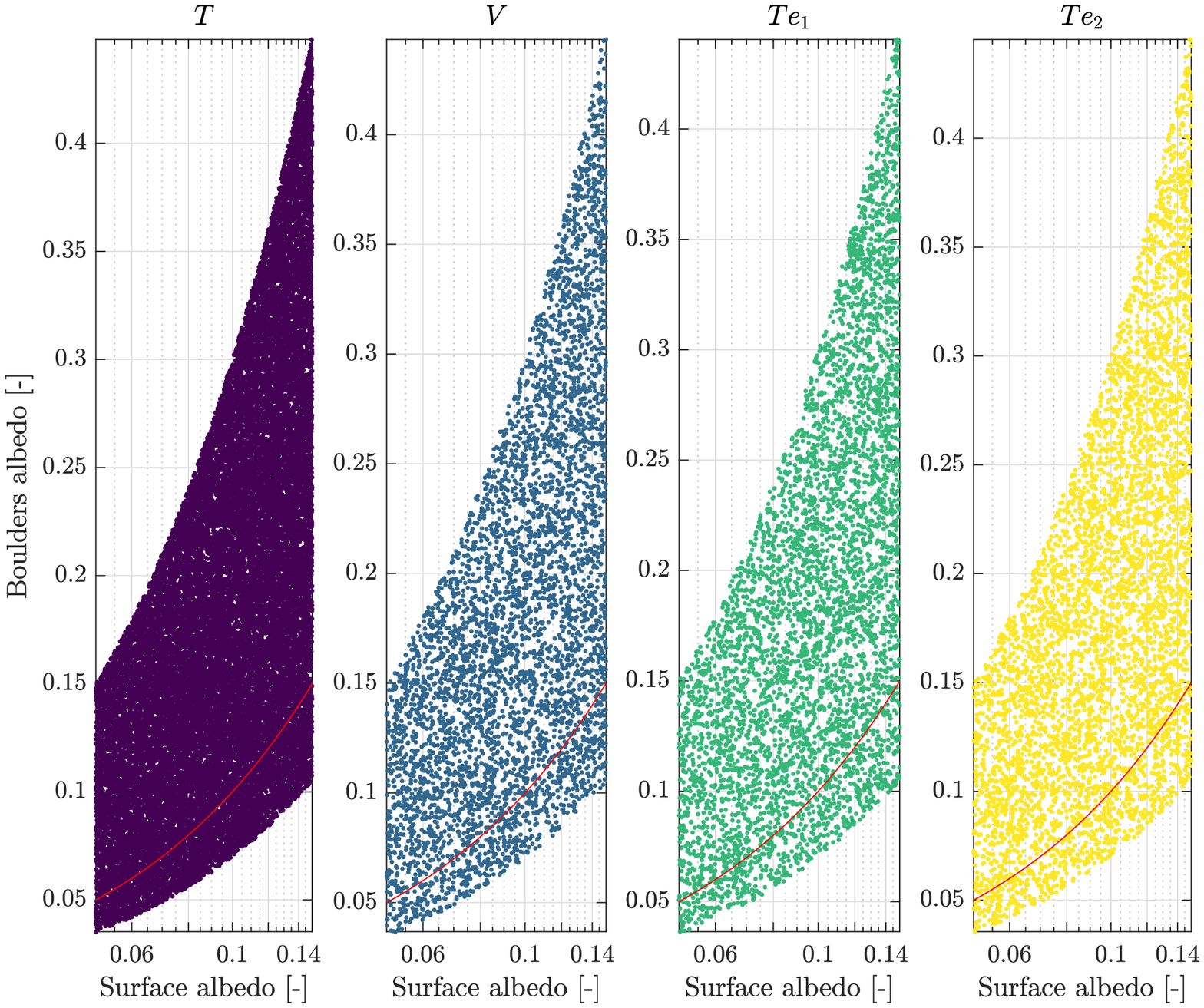}
	\caption{Boulders albedo as function of the surface albedo for the various samples of the splits of $DS_1$. The red line represents the albedo equality condition.}
	\label{fig:DS1_albedo_1}
\end{figure}

\begin{figure}[ht]
	\centering
	\includegraphics[width=0.85\textwidth]{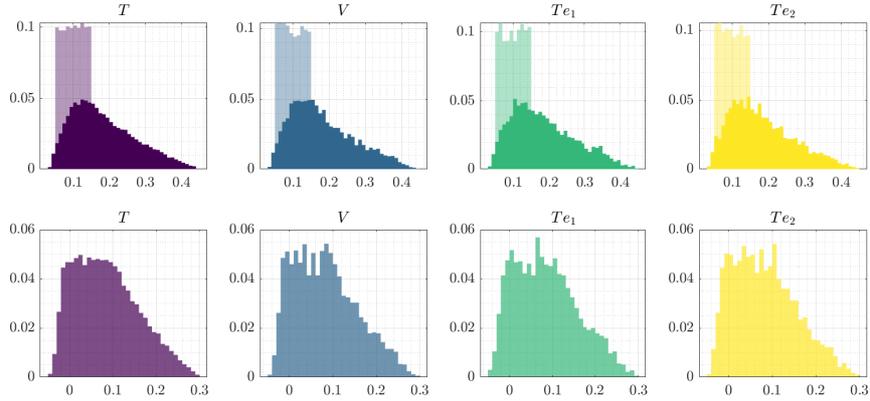}
	\caption{Histograms of surface (transparent) and boulder (solid) albedo's (top) and their difference (bottom) in $DS_1$. Bin-width used: 0.01.}
	\label{fig:DS1_albedo_2}
\end{figure}

In Figure \ref{fig:DS1_XYZ} the $X,Y,Z$ coordinates of the camera are represented in the Blender reference frame. In Figure \ref{fig:DS1_histograms} various histograms are illustrated for the $X,Y,Z$ coordinates, boresight rotation angle $\theta_b$, range $\rho$, phase angle $\Psi$, and Sun's intensity $I$. In particular, it is interesting to note the different distributions of $\Psi$ between the $Te_2$ split and $Tr$, $V$, and $Te_1$ splits. These distributions reflect the balanced nature of the latter datasets. Finally, the relationship between boulder's and surface albedo is illustrated in Figure \ref{fig:DS1_albedo_1} while in Figure  \ref{fig:DS1_albedo_2} their histograms are represented.

\clearpage
\newpage

\clearpage
\newpage

\FloatBarrier
\section{Statistical properties of $DS_2$}
The $DS_2$ dataset has been designed for different IP applications. Among these, the authors recognize boulders identification, segmentation, and navigation. The main statistical properties of $DS_2$ are illustrated in detail. Note that all histograms have been plotted with the value of the relative probability on the y-axis corresponding to each bin. 

\begin{figure}[ht]
    \begin{minipage}{.4\linewidth}
        \centering
        \includegraphics[scale=.4]{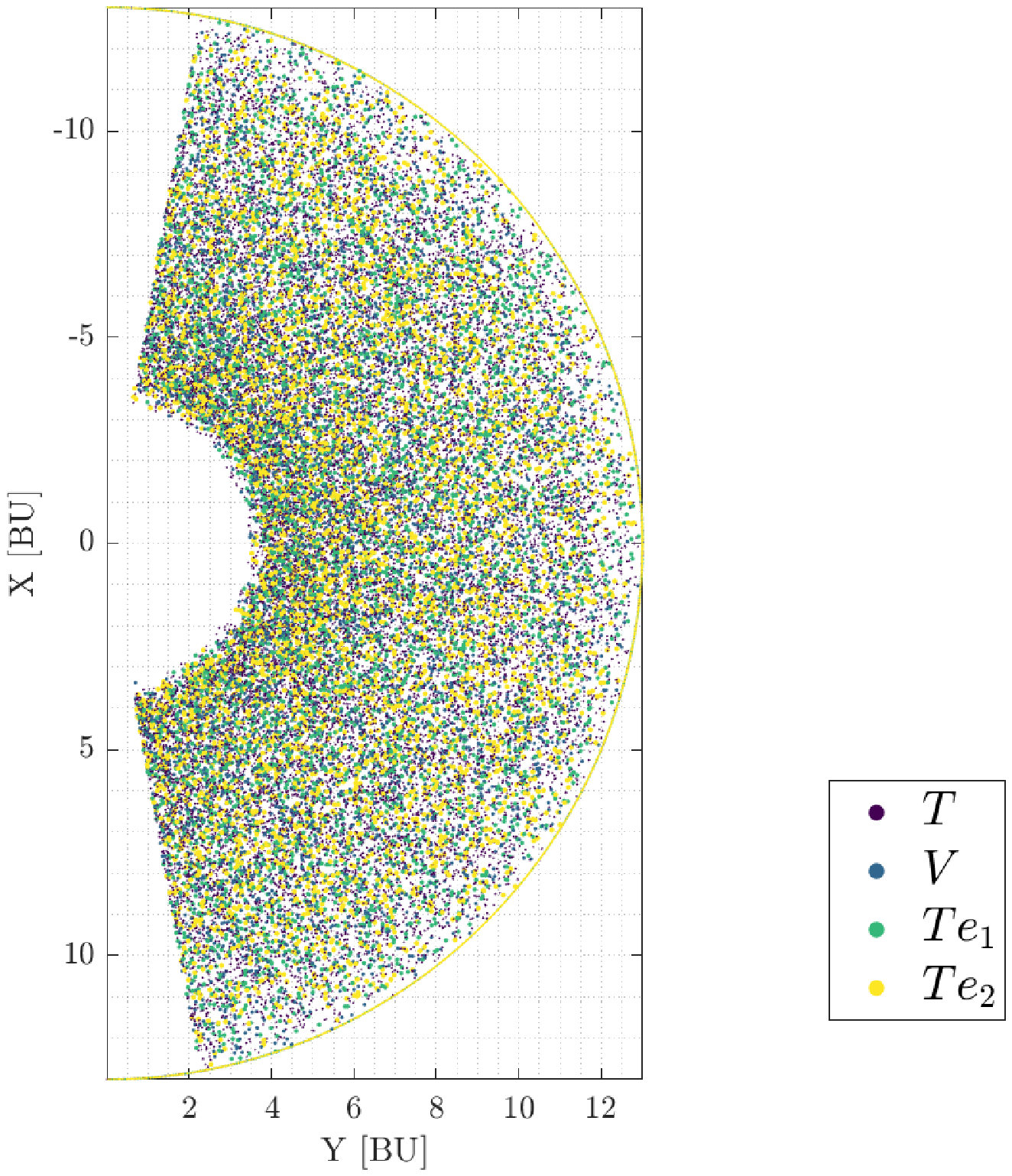}
    \end{minipage}%
    \begin{minipage}{.4\linewidth}
        \centering
        \includegraphics[scale=.4]{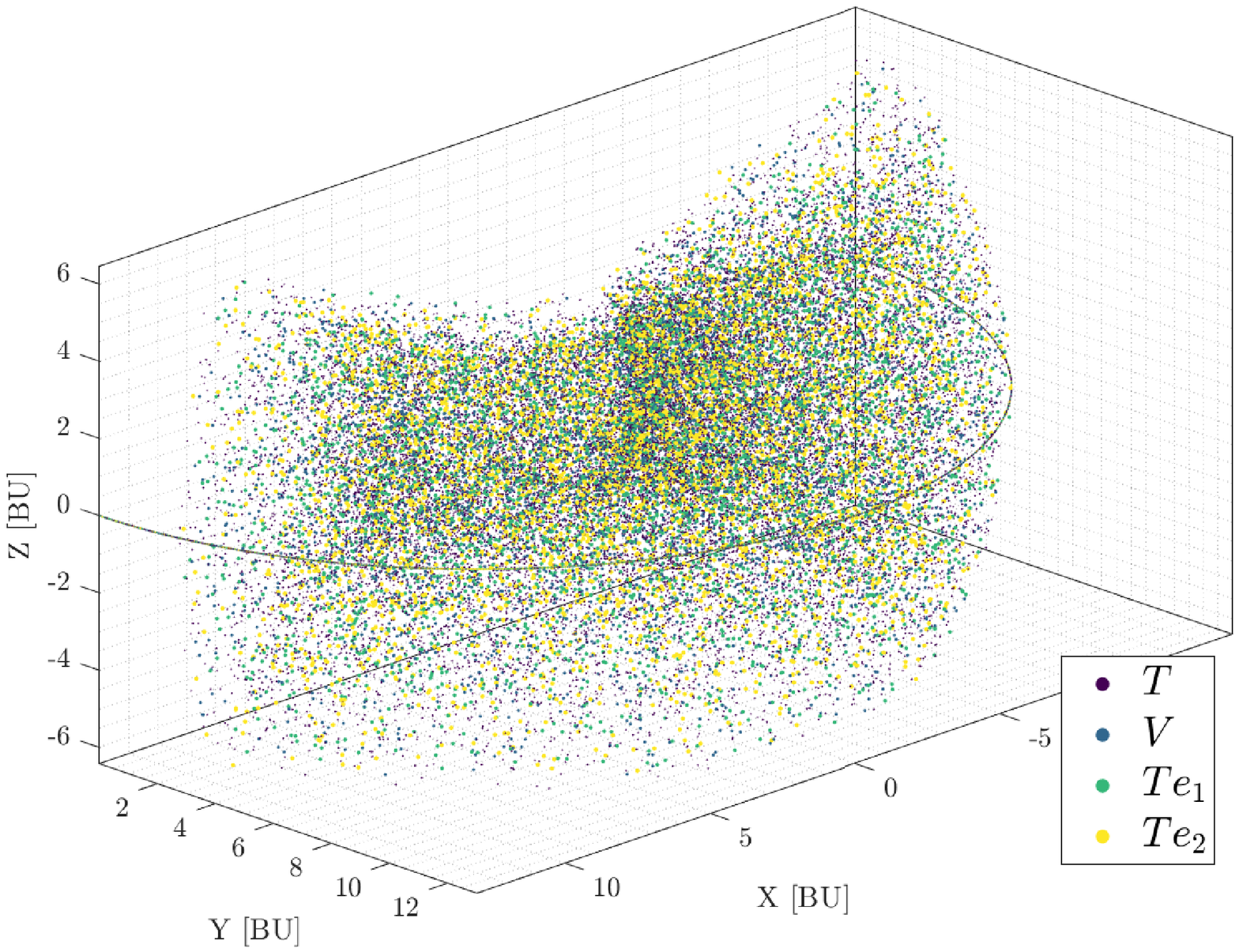}
    \end{minipage}\par\medskip
\caption{Cloud of points of positions and Sun's direction used in the various splits of $DS_2$ in the Blender reference frame. Sun direction varies only in the equatorial plane.}
\label{fig:DS2_XYZ}
\end{figure}

\begin{figure}[ht]
	\centering
	\includegraphics[width=1\textwidth]{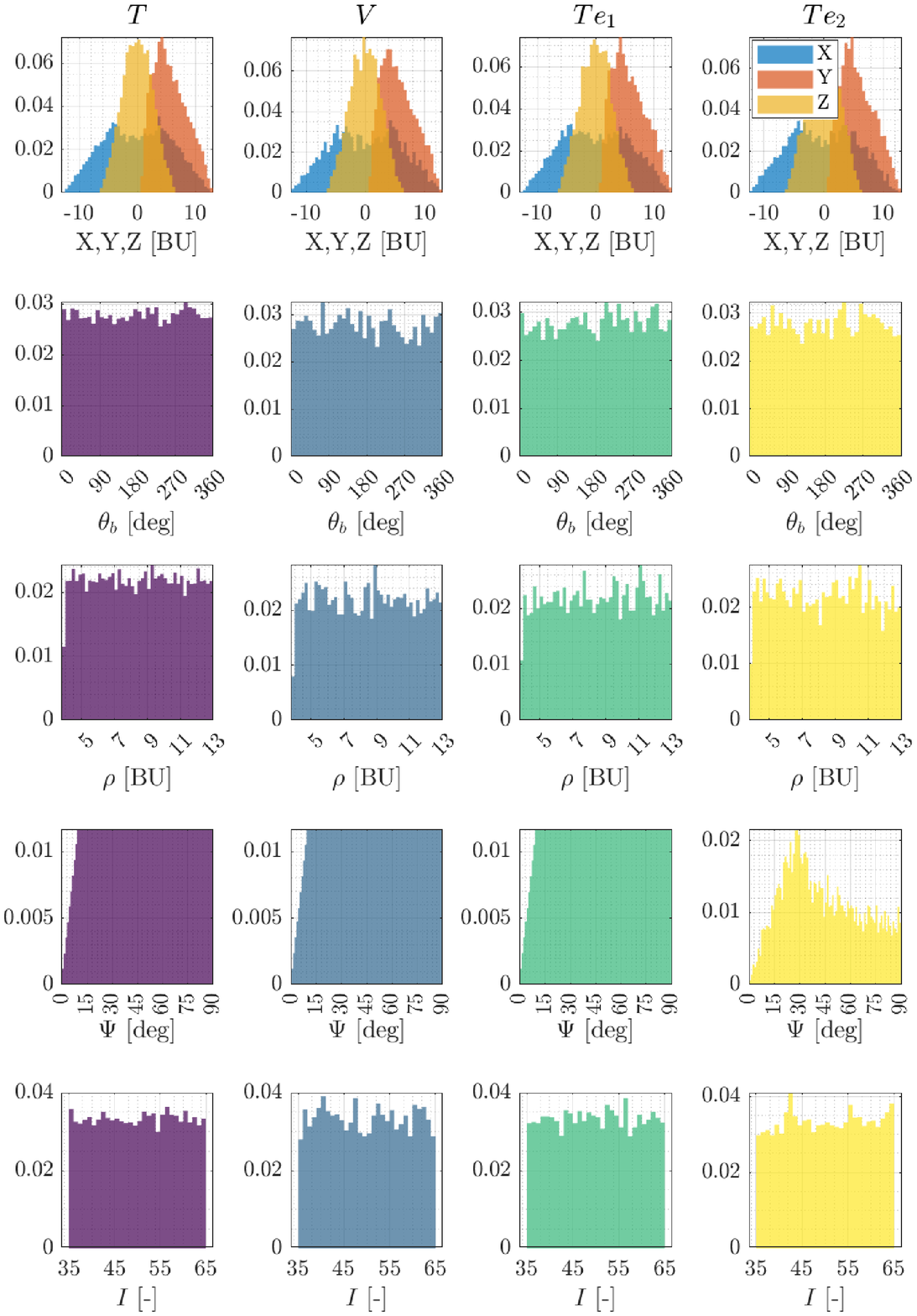}
	\caption{Histograms of properties of the various splits of $DS_2$. From top to bottom: Cartesian coordinates, boresight rotation, range, phase angle, and Sun's intensity. Bin-width used from top to bottom: 0.5, 10 deg, 0.2, 1 deg, 1.}
	\label{fig:DS2_histograms}
\end{figure}

\begin{figure}[ht]
	\centering
	\includegraphics[width=0.7\textwidth]{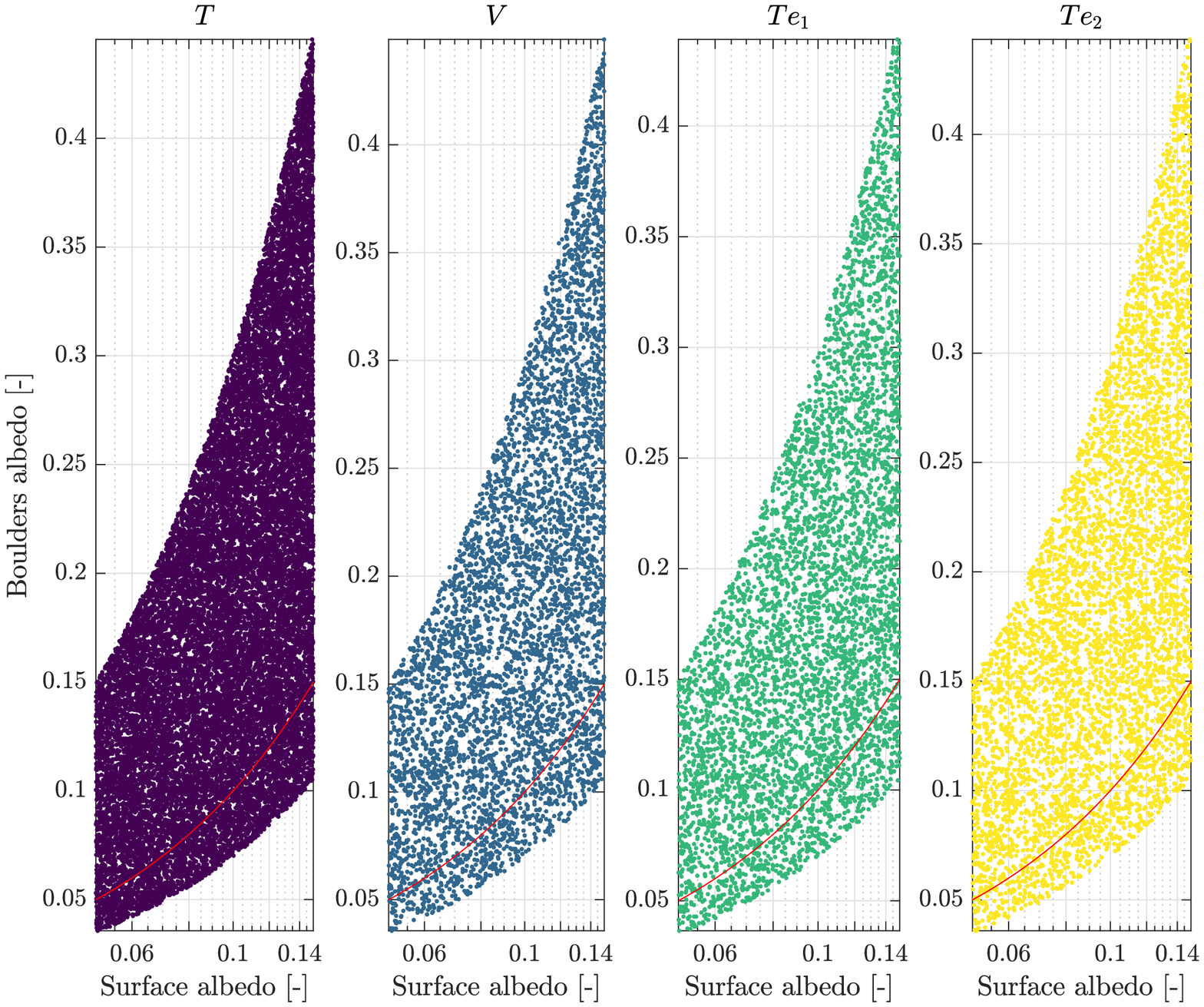}
	\caption{Boulders albedo as function of the surface albedo for the various samples of the splits of $DS_2$. The red line represents the albedo equality condition.}
	\label{fig:DS2_albedo_1}
\end{figure}

\begin{figure}[ht]
	\centering
	\includegraphics[width=0.85\textwidth]{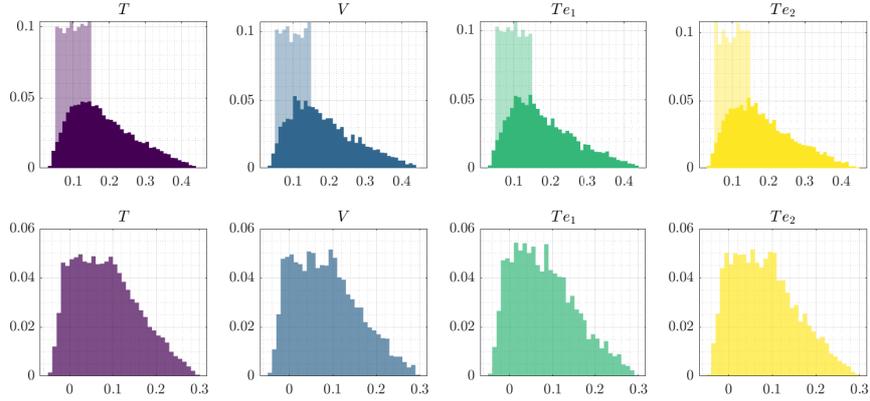}
	\caption{Histograms of surface (transparent) and boulder (solid) albedo's (top) and their difference (bottom) in $DS_2$. Bin-width used: 0.01.}
	\label{fig:DS2_albedo_2}
\end{figure}

In Figure \ref{fig:DS2_XYZ} the $X,Y,Z$ coordinates of the camera are represented in the Blender reference frame. In Figure \ref{fig:DS2_histograms} various histograms are illustrated for the $X,Y,Z$ coordinates, boresight rotation angle $\theta_b$, range $\rho$, phase angle $\Psi$, and Sun's intensity $I$. In particular, it is interesting to note the different distributions of $\Psi$ between the $Te_2$ split and $Tr$, $V$, and $Te_1$ splits. These distributions reflect the balanced nature of the latter datasets. Finally, the relationship between boulder's and surface albedo is illustrated in Figure \ref{fig:DS2_albedo_1} while in Figure \ref{fig:DS2_albedo_2} their histograms are represented.

\clearpage
\newpage

\clearpage
\newpage

\FloatBarrier

\section{Conclusions}
In this work the authors have presented the Blender setup and statistical characterization of two datasets that can be used for various IP tasks about small bodies. The datasets have been first used in \cite{pugliatti2022_siw} and are made publicly available at \cite{DatasetZenodoBoulders}.

\section*{Acknowledgment}
The authors would like to acknowledge the funding received from the European Union’s Horizon 2020 research and innovation programme under the Marie Skłodowska-Curie grant agreement No 813644.

\printbibliography

@article{song2022deep,
    title={Deep learning-based spacecraft relative navigation methods: A survey},
    author={Song, Jianing and Rondao, Duarte and Aouf, Nabil},
    journal={Acta Astronautica},
    volume={191},
    pages={22--40},
    year={2022},
    publisher={Elsevier}
}

@INPROCEEDINGS{pugliatti2022_siw,
    title = {Boulders identification on small bodies under varying illumination conditions},
    author = {Mattia Pugliatti and Francesco Topputo},
    booktitle={3rd Space Imaging Workshop, Atlanta, GA},
    number={SIW22-09},
    year = {2022},
    month={10},
    pages = {1--12}
}

@INPROCEEDINGS{pentila2021scattering,
    title = {Realistic visualization of solar system small bodies using Blender ray tracing software},
    author = {Antti Penttilä and Mario F. Palos and Tomas Kohout},
    booktitle={European Planetary Science Congress, Wien, Austria},
    number={https://doi.org/10.5194/epsc2021-791},
    year = {2021},
    month={9},
    pages = {1}
}

@dataset{DatasetZenodoBoulders,
    author       = {Mattia Pugliatti and Francesco Topputo},
    title        = {{DOORS: Dataset fOr bOuldeRs Segmentation}},
    month        = {9},
    year         = {2022},
    publisher    = {Zenodo},
    version      = {1.0},
    doi          = {10.5281/zenodo.7107409},
    url          = {https://doi.org/10.5281/zenodo.7107409}, note = {Sept 2022, Zenodo, V1.0, doi: 10.5281/zenodo.7107409}
}
\end{document}